\documentclass[abstract,a4paper]{scrartcl}

\usepackage[english]{babel}
\usepackage{amssymb,latexsym,amsmath,stmaryrd}
\usepackage[usenames,dvipsnames]{color}
\usepackage{graphicx}
\usepackage{calc}
\usepackage{xspace}
\usepackage{url}
\usepackage[ruled,noline,noend,linesnumbered]{algorithm2e}
\usepackage[normalem]{ulem}
\usepackage{paralist}
\usepackage{comment}
\usepackage{multirow}

\usepackage{amsthm}
\usepackage[numbers]{natbib}
\newcommand{\institute}[1]{\publishers{\small #1}}
\newcommand{\email}[1]{\textsf{#1}}

\includecomment{WithAppendix}

\SetAlFnt{\small}

\newcommand{\ignore}[1]{}

\newcommand{\as}[1]%
{\marginpar{\textcolor{red}{\textsc{Andreas}}}\textcolor{red}{[[#1]]}}
\newcommand{\pjs}[1]%
{\marginpar{\textcolor{OliveGreen}{\textsc{Peter}}}\textcolor{OliveGreen}{[[#1]]}}
\newcommand{\tf}[1]%
{\marginpar{\textcolor{blue}{\textsc{Thibaut}}}\textcolor{blue}{[[#1]]}}

\newcommand{\cp}{\textsc{Cp}\xspace}

\newcommand{\lcg}{\textsc{Lcg}\xspace}

\newcommand{\rcpsp}{\textsc{Rcpsp}\xspace}

\newcommand{\sat}{\textsc{Sat}\xspace}
\newcommand{\ttef}{\textsc{TtEf}\xspace}
\newcommand{\vsids}{\textsc{Vsids}\xspace}
\newcommand{\hs}{\textsc{HotStart}\xspace}
\newcommand{\hr}{\textsc{HotRestart}\xspace}
\newcommand{\ttProp}{\textsf{tt}\xspace}
\newcommand{\ttefCheck}{\textsf{ttef(c)}\xspace}
\newcommand{\ttefProp}{\textsf{ttef}\xspace}

\newcommand{\FontGlobals}[1]{\texttt{#1}\xspace}
\newcommand{\cumu}{\FontGlobals{cumulative}}

\newcommand{\OO}{\mathcal{O}\xspace}   
\newcommand{\VV}{\mathcal{V}\xspace}   

\newcommand{\lit}[1]{\llbracket #1 \rrbracket}

\DeclareMathOperator{\true}{\mathit{true}}

\newcommand{\lb}[1]{lb(#1)}
\newcommand{\ub}[1]{ub(#1)}

\newcommand{\energy}[1]{e_{#1}}
\newcommand{\est}[1]{est_{#1}}
\newcommand{\lst}[1]{lst_{#1}}
\newcommand{\ect}[1]{ect_{#1}}
\newcommand{\lct}[1]{lct_{#1}}
\newcommand{\eEF}[1]{\energy{#1}^{E\!F}}
\newcommand{\pEF}[1]{p^{E\!F}_{#1}}
\newcommand{\eTT}[1]{\energy{#1}^{TT}}
\newcommand{\pTT}[1]{p^{TT}_{#1}}
\newcommand{\lstEF}[1]{lst^{E\!F}_{#1}}
\newcommand{\setV}{\VV}
\newcommand{\setEF}{\VV^{E\!F}}
\newcommand{\ttEn}{ttEn}
\newcommand{\ttAfter}[1]{ttAfter[#1]}
\newcommand{\rsEn}{rsEn}

\newcommand{\benchAT}{\textsc{AT}\xspace}
\newcommand{\benchBL}{\textsc{BL}\xspace}
\newcommand{\benchJ}[1]{\textsc{j#1}\xspace}
\newcommand{\benchKSDD}[1]{\textsc{KSD#1\textunderscore d}\xspace}
\newcommand{\benchPack}{\textsc{Pack}\xspace}
\newcommand{\benchPackD}{\textsc{Pack\textunderscore d}\xspace}

\newcommand{\ie}{\emph{i.e.},\xspace}

\theoremstyle{plain}

\newtheorem{proposition}{Proposition}[section]

\theoremstyle{definition}

\newtheorem{example}{Example}[section]

\begin{document}

\title{Explaining Time-Table-Edge-Finding Propagation for the Cumulative Resource Constraint}

\author{
   Andreas Schutt \and
   Thibaut Feydy \and
   Peter J. Stuckey
}

\institute{
	Optimisation Research Group, National ICT Australia, and Department of Computing and Information Systems,
	The~University~of~Melbourne, Victoria 3010, Australia\\
	\email{\{andreas.schutt,thibaut.feydy,peter.stuckey\}@nicta.com.au}\\
}

\date{}
\maketitle

\begin{abstract}
	Cumulative resource constraints can model scarce resources in scheduling problems or a dimension in packing and cutting problems.
	In order to efficiently solve such problems with a constraint programming solver, it is important to have strong and fast propagators for cumulative resource constraints.
	One such propagator is the recently developed time-table-edge-finding 
	propagator, which considers the current resource profile during the edge-finding propagation.
	Recently, lazy clause generation solvers, \ie constraint programming solvers
    incorporating nogood learning, have proved to be excellent at
	solving scheduling and cutting problems.
	For such solvers, concise and accurate explanations of the reasons
	for propagation are essential for strong nogood learning.
	In this paper, 
	we develop the first explaining version of time-table-edge-finding
propagation and show preliminary results on 
resource-constrained project scheduling problems from various standard
benchmark suites.
	On the standard benchmark suite PSPLib, we were able to close one open instance and to improve the lower bound of about 60\% of the remaining open instances. Moreover, 6 of those instances were closed.
\end{abstract}

\section{Introduction}

A cumulative resource constraint models the relationship between a scarce
resource and activities requiring some part of the resource capacity for their execution.
Resources can be workers, processors, water, electricity, or, even, a
dimension in a packing and cutting problem.  
Due to its relevance in many industrial scheduling and placement problems, 
it is important to have strong and fast propagation techniques in
constraint programming (\cp) solvers that detect inconsistencies early and remove many invalid values from the domains of the variables involved.  
Moreover, when using \cp
solvers that incorporate ``fine-grained'' nogood learning it is also important that
each inconsistency and each value removal from a domain is explained in
such a way that the full strength of nogood learning is exploited.

In this paper, we consider \emph{renewable} resources, \ie resources with a
constant resource capacity over time, and \emph{non-preemptive} activities,
\ie whose execution cannot be interrupted, with fixed processing times and
resource usages.
In this work, we develop explanations for the 
time-table-edge-finding (\ttef) propagator~\cite{Vilim:11} for use 
in lazy clause generation (\lcg) solvers~\cite{Ohrimenko:SC:09,Feydy:S:09}.

\begin{figure}[tp]
\begin{minipage}[b]{0.45\linewidth}
	\centering
	\def\svgwidth{5cm} 
	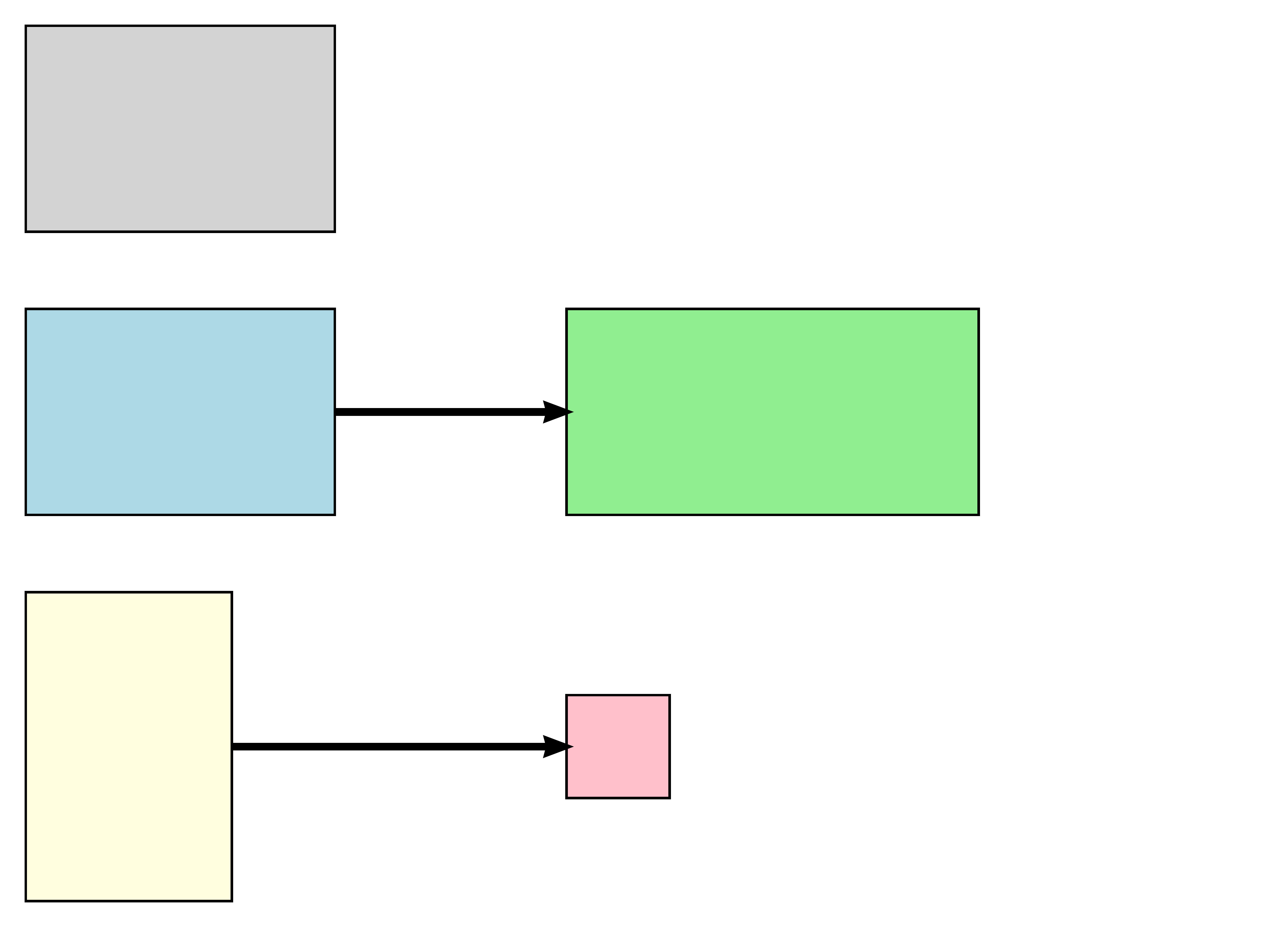
	\caption{Five activities with precedence relations.}
	\label{fig:ex_intro_aon}
\end{minipage}
	\hfill
\begin{minipage}[b]{0.45\linewidth}
	\centering
	\def\svgwidth{5cm} 
	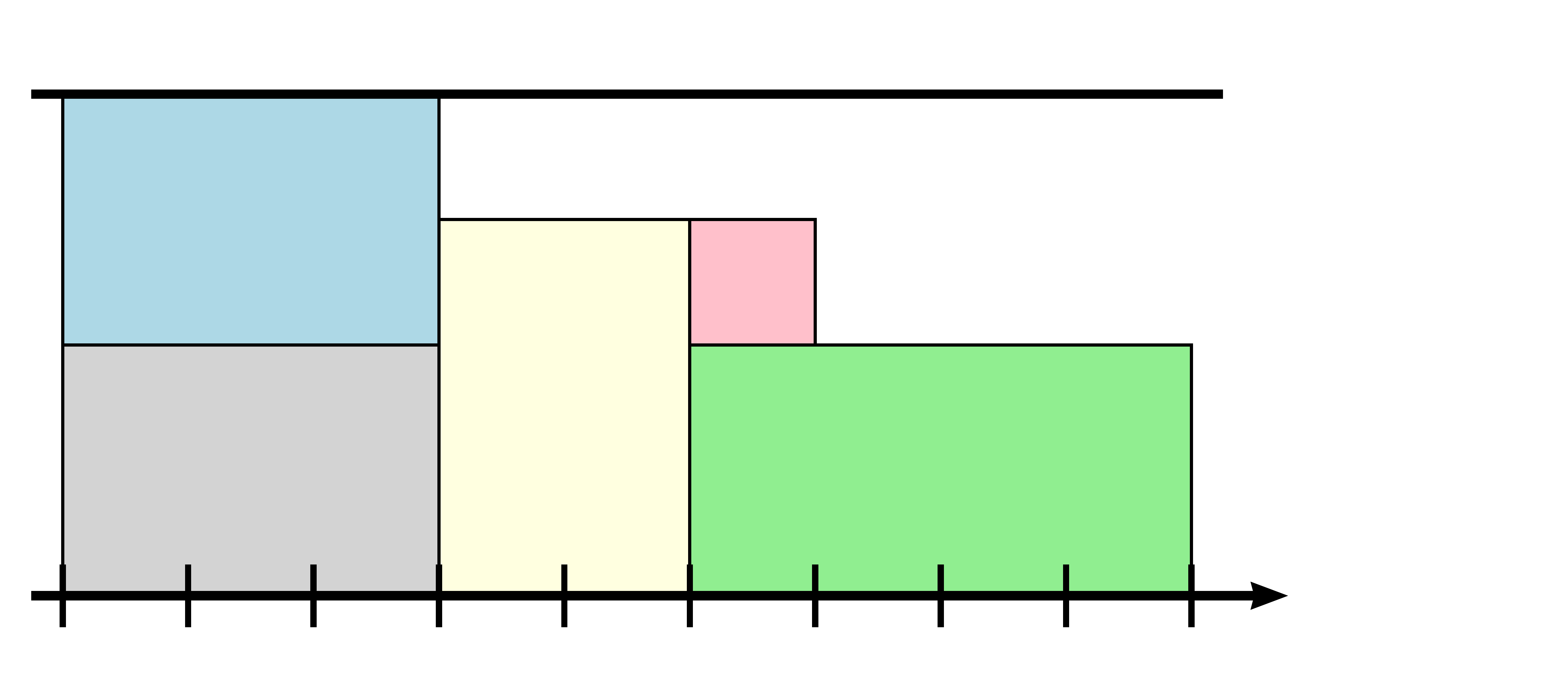 
	\caption{A possible schedule of the activities.}
	\label{fig:ex_intro_sched}
\end{minipage}
\end{figure}

\begin{example}
	Consider a simple cumulative resource scheduling problem.
	There are 5 activities A, B, C, D, and E to be executed before time period 10.
	The activities have processing times 3, 3, 2, 4, and 1, respectively,
	with each activity requiring 2, 2, 3, 2, and 1 units of resource,
        respectively. There is a resource capacity of 4.  
	Assume further that there are precedence constraints: activity B must
	finish before activity D begins, written	$B \ll D$,
	and similarly $C \ll E$.
	Figure~\ref{fig:ex_intro_aon} shows the five activities and
        precedence relations, while Fig.~\ref{fig:ex_intro_sched} shows a possible schedule, where the
        start times are: 0, 0, 3, 5, and 5 respectively.
\end{example}

In \cp solvers, a cumulative resource constraint can be modelled by a decomposition or, more successfully, by the global constraint \cumu~\cite{Aggoun:B:93}.
Since the introduction of this global constraint, a great deal of research has investigated stronger and faster propagation techniques.
These include time-table~\cite{Aggoun:B:93}, (extended)
edge-finding~\cite{Nuijten:94,Vilim:09}, not-first/not-last~\cite{Nuijten:94,Schutt:W:10}, and energetic-reasoning propagation~\cite{Baptiste:LP:00,Baptiste:LPN:01}.
Time-table propagation is usually superior for \emph{highly disjunctive} problems, 
\ie in which only some activities can run concurrently, 
while (extended) edge-finding, not-first/not-last, and energetic reasoning are more appropriate for \emph{highly cumulative} problems, 
\ie in which many activities can run concurrently.\cite{Baptiste:LP:00}
The reader is referred to~\cite{Baptiste:LPN:01} 
for a detailed comparison of these techniques.

Vilim~\cite{Vilim:11} recently developed \ttef propagation which combines the
time-table and (extended) edge-finding propagation in order to perform
stronger propagation while having a low runtime overhead.
Vilim~\cite{Vilim:11} shows that on a range of highly disjunctive open resource-constrained
project scheduling problems from the well-established benchmark
library PSPLib,\footnote{See \url{http://129.187.106.231/psplib/}.} 
\ttef propagation can generate lower bounds on the project deadline
(\emph{makespan}) that are superior to those found by previous methods.
He uses a \cp solver without nogood learning.  
This result, and the success of
\lcg on such problems, motivated us to study whether an explaining version of
this propagation yields an improvement in performance for \lcg solvers.

In general, nogood learning is a resolution step that infers redundant constraints, called \emph{nogoods}, given an inconsistent solution state.
These nogoods are permanently or temporarily added to the initial constraint system in order to reduce the search space and/or to guide the search.
Moreover, they can be used to short circuit propagation.
How this resolution step is performed is dependent on the underlying system.

\lcg solvers employ a ``fine-grained'' nogood learning system that mimics the learning of modern Boolean satisfiability (\sat) solvers (see e.g. \cite{Moskewicz:MZZM:01}).
In order to create a strong nogood, 
it is necessary that each inconsistency and value removal is explained 
concisely and in the most general way possible.
For \lcg solvers, we have previously developed explanations for time-table and  (extended) edge-finding propagation~\cite{Schutt:FSW:11}.
Moreover, for time-table propagation we have also considered the case when processing times, resource usages, and resource capacity are variable~\cite{Schutt:11}.
Explanations for the time-table propagator were successfully applied on
resource-constraint project scheduling
problems~\cite{Schutt:FSW:11,Schutt:FSW:12} and carpet
cutting~\cite{Schutt:SV:11} where in both cases the state-of-the-art of
exact solution methods were substantially improved.
The explanations defined here 
are similar to the step-wise ones for the (extended) edge-finding propagation in~\cite{Schutt:FSW:11}, but there we do not consider the resource profile and are more complex.
Moreover, the proposed explanations for edge-finding propagation in~\cite{Schutt:FSW:11} has never been implemented.

Explanations for the propagation of the cumulative constraint have also been proposed for the PaLM~\cite{Jussien:B:00,Jussien:03} and SCIP~\cite{Achterberg:09,Berthold:HLMS:10,Heinz:S:11} frameworks. 
In the PaLM framework, explanations are only considered for time-table propagation, while the SCIP framework additionally provides explanations for energetic reasoning propagation and a restricted version of edge-finding propagation.
Neither framework consider bounds widening in order to generalise these explanations as we do in this paper.
Other related works include \cite{Vilim:05}, which presents explanations for
different propagation techniques for problems only involving disjunctive
resources, \ie cumulative resources with unary resource capacity, and
generalised nogoods~\cite{Katsirelos:B:05}.
A detailed comparison of 
explanations for the propagation of cumulative resource constraints 
in \lcg solvers can be found in~\cite{Schutt:11}.

In this paper we develop explanations for the \ttef cumulative propagator 
in \lcg solvers. 
The explaining \ttef propagation is then compared with the explaining time-table propagation from~\cite{Schutt:FSW:11} in the \lcg solver on \rcpsp using the reengineered \lcg solver~\cite{Feydy:S:09} which was also used for the experiments presented in~\cite{Schutt:FSW:11}.

\section{Cumulative Resource Scheduling}

In cumulative resource scheduling, a set of (non-preemptive)
activities~$\setV$ and one cumulative resource with a (constant) resource
capacity~$R$ is given where an \emph{activity}~$i$ is specified by its
\emph{start time}~$S_i$, its \emph{processing time}~$p_i$, its
\emph{resource usage}~$r_i$, and its \emph{energy}~$\energy{i} := p_i \cdot
r_i$.
In this paper we assume each $S_i$ 
is an integer variable and all others are assumed to be
integer constants.  
Further, we define $\est{i}$ ($\ect{i}$) and
$\lst{i}$ ($\lct{i}$) as the \emph{earliest} and \emph{latest} start
(completion) time of~$i$.  

In this setting. the cumulative resource scheduling problem is defined as 
a constraint satisfaction problem that is characterised by the set of activities~$\setV$ 
and a cumulative resource with resource capacity~$R$. The goal is to find a solution that assigns values from the domain to the start time variables~$S_i$ ($i\in \setV$), so that the following conditions are satisfied.
\begin{align*}
	&\est{i} \leq S_i \leq \lst{i}, & \forall i \in \setV\\
	&\sum_{i\in \setV: \tau \in [S_i, S_i + p_i)} r_i \leq R & \forall \tau
\end{align*}
where $\tau$ ranges over the time periods considered.
Note that this problem is NP-hard~\cite{Baptiste:LN:99}.

We shall tackle problems including cumulative resource scheduling using \cp with nogood learning.
In a \cp solver, each variable $S_i, i \in \setV$ has an initial domain of possible values $D^0(S_i)$ which is initially $[\est{i}, \lst{i}]$.
The solver maintains a current domain $D$ for all variables.
\cp search interleaves propagation with search.
The constraints are represented by propagators that, given the current domain
$D$, creates a new smaller domain $D'$ by eliminating infeasible values.
The current \emph{lower} and \emph{upper bound} of the domain $D(S_i)$ are denoted by $\lb{S_i}$ and $\ub{S_i}$, respectively.
For more details on \cp see e.g.~\cite{Schulte:S:08}.

For a learning solver we also represent the domain of each variable $S_i$ using Boolean variables $\lit{S_i \leq v}, \est{i} \leq v < \lst{i}$. 
These are used to track the reasons for propagation and generate nogoods.
For more details see~\cite{Ohrimenko:SC:09}.  
We use the notation $\lit{v \leq S_i}, \est{i} < v \leq \lst{i}$ 
as shorthand for $\neg \lit{S_i \leq v - 1}$, and treat $\lit{v \leq S_i}, v
\leq \est{i}$ and
$\lit{S_i \leq v}, v \geq \lst{i}$ as synonyms for $\true$.
Propagators in a learning solver must explain each reduction in the domain by building a clausal explanation using these Boolean variables.

Optimisation problems are typically solved in \cp via branch and bound. 
Given an objective $obj$ which is to be minimised, when a solution is found with objective value $o$, a new constraint $obj < o$ is posted to enforce that we only look for better solutions in the subsequent search.

\section{TTEF Propagation}

In this section we develop explanations for \ttef propagation.
For a more detailed description about \ttef propagation the reader is referred to~\cite{Vilim:11}.

\begin{figure}[tp]
	\centering
	\def\svgwidth{6cm} 
	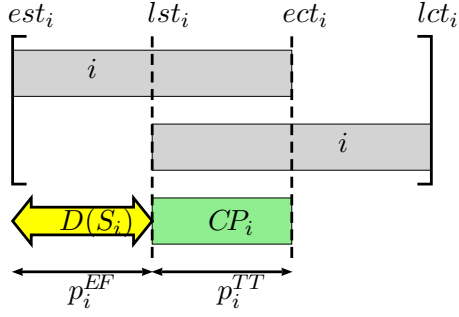 
	\caption{A diagram illustrating an activity $i$ when started at $\est{i}$ or $\lst{i}$, and its possible range of start
	times, as well as the compulsory part $C\!P_i$, and the fixed and free parts
	of the processing time.\label{fig:task}}
\end{figure}

\ttef propagation splits the treatment of activities into a fixed and free part.
The former results from the activities' compulsory part whereas the latter is the remainder.
The fixed part of an activity~$i$ is characterised by the length of its compulsory part~$\pTT{i} := \max(0, \ect{i} - \lst{i})$ and its fixed energy $\eTT{i} := r_i \cdot \pTT{i}$. 
The free part has a processing time~$\pEF{i} := p_i - \pTT{i}$ and a free energy of~$\eEF{i} := \energy{i} - \eEF{i}$.
Let $\setEF$ be the set of activities with a non-empty free part $\{i \in \setV \mid \pEF{i} > 0\}$.
An illustration of this is shown in Figure~\ref{fig:task}.

\ttef propagation reasons about the energy available from the resource and energy required for the execution of activities in specific time windows.
The start and end times of these windows are determined by the earliest start and the latest completion times of activities~$i \in \setEF$.
These time windows $[begin, end)$ are characterised by the so-called \emph{task intervals}~$\setEF(a,b) := \{ i \in \setEF \mid \est{a} \leq \est{i} \wedge \lct{i} \leq \lct{b} \}$ where ${a, b} \in \setEF$, $begin := \est{a}$, and $end := \lct{b}$. 

It is not only the free energy of activities in the task interval~$\setEF(a,b)$ that is considered, but also the energy resulting from the compulsory parts in the time window~$[\est{b}, \lct{b})$.
This energy is defined by $\ttEn(a, b) := \ttAfter{\est{a}} - \ttAfter{\lct{b}}$ where $\ttAfter{\tau} := \sum_{t \geq \tau} \sum_{i \in \setV : \lst{i} \leq t < \ect{i}} r_i$.

Furthermore, we also consider activities~$i \in \setV \setminus \setEF(a, b)$ in which a portion of their free part must be run within the time window as described in~\cite{Vilim:11}.
Let $\lstEF{i}$ be the latest start time of the free part of an activity, \ie $\lct{i} - \pEF{i}$.
Then activity~$i$'s free part consumes at least $r_i \cdot (\lct{b} - \lstEF{i})$ energy units in~$[\est{a}, \lct{b})$ if $\est{a} \leq \est{i}$ and $\lstEF{i} < \lct{b}$.
We define the energy contributed by such activities by $\rsEn(a, b) := \sum_{i \in \setV \setminus \setEF(a, b) : \est{a} \leq \est{i}} max(0, \lct{b} - \lstEF{i})$.

In summary, \ttef propagation considers three ways in which an activity~$i$ can contribute to energy consumption within a time window determined by a task interval~$\setEF(a, b)$.
First, the free parts that must fully be executed in the time window; second, the compulsory parts that must lies in the time window; and third, some free parts that must partially be run in the time window.
Thus, the considered length of an activity~$i$ is
\begin{equation*}
	p_i(a, b) := 
	\left\{ 
	\begin{aligned}
	p_i & \quad i \in \setEF(a,b)\\
	\max(0, \lct{b} - \lst{i}) & \quad i \notin \setEF(a,b) \wedge \est{a} \leq \est{i}\\
	\max(0, \min(\lct{b}, \ect{i}) - \max(\est{a}, \lst{i})) & \quad others
	\end{aligned}
	\right.
\end{equation*}
The considered energy consumption is $\energy{i}(a,b) := r_i \cdot p_i(a,b)$ in the time window.

\subsection{Explanation for the TTEF Consistency Check}
\label{ssec:ttefc}

The consistency check is one part of \ttef propagation that checks whether there
is a resource overload in any task interval.
\begin{proposition}[Consistency Check]
	The cumulative resource scheduling problem is inconsistent if
	\begin{equation}
	R \cdot (\lct{b} - \est{a}) - energy(a,b) < 0 \to \bot
	\end{equation}
	where $energy(a,b) := \sum_{i\in \setEF(a, b)} \eEF{i} + \ttEn(a, b) + \rsEn(a, b)$.
\end{proposition}
This check can be done in $\OO(l^2 + n)$ runtime, where $l = |\setEF|$, if the resource profile is given.
\begin{WithAppendix}
The corresponding algorithm is shown in Alg.~\ref{alg:ttef_check} in App.~\ref{app:sec:ttef:alg}.
\end{WithAppendix}

A na\"{\i}ve explanation for a resource overload in the time window $[\est{a}, \lct{b})$ only considers the current bounds on activities' start times~$S_i$.
\begin{equation*}
\bigwedge_{i \in \setV : p_i(a, b) > 0} 
	\lit{\est{i} \leq S_i} \wedge \lit{S_i \leq \lst{i}} \to \bot
\end{equation*}
However, we can easily generalise this explanation by only ensuring that at least $p_i(a,b)$ time units are executed in the time window.
This results in the following explanation.
\begin{equation*}
\bigwedge_{i \in \setV : p_i(a, b) > 0} 
	\lit{\est{a} + p_i(a,b) - p_i \leq S_i} \wedge \lit{S_i \leq \lct{b} - p_i(a,b)} \to \bot
\end{equation*}
Note that this explanation expresses a resource overload with respect to energetic reasoning propagation which is more general than \ttef.

Let $\Delta := energy(a, b) - R \cdot (\lct{b} - \est{a}) - 1$. 
If $\Delta > 0$ then the resource overload has extra energy. We can use this
extra energy to further generalise the explanation, by reducing the energy
required to appear in the time window by up to $\Delta$. 
For example, if $r_i \leq \Delta$ then the lower and upper bound on 
$S_i$ can simultaneously be decreased and increased by a total amount in 
$\{1, 2, ..., \min(\lfloor \Delta / r_i \rfloor, p_i(a,b))\}$ units without resolving the overload. 
If $r_i \cdot p_i(a,b) \leq \Delta$ then we can remove activity $i$ 
completely from the explanation.
In a greedy manner, we try to maximally widen the bounds of activities
$i$ where $p_i(a,b) > 0$, first considering activities with non-empty free parts. 
If $\Delta_i$ denotes the time units of the widening then it holds 
$p_i(a,b) \geq \Delta_i \geq 0$ and 
$\sum_{i \in \setV : p_i(a, b) > 0} \Delta_i\cdot r_i \leq \Delta$ and we create the following explanation.
\begin{equation*}
\bigwedge_{i \in \setV : p_i(a, b) - \Delta_i > 0} 
	\lit{\est{a} + p_i(a,b) - p_i - \Delta_i \leq S_i} \wedge \lit{S_i \leq \lct{b} - p_i(a,b) + \Delta_i} \to \bot
\end{equation*}

The last generalisation mechanism can be performed in different ways, e.g. we could widen the bounds of activities that were involved in many recent conflicts.
Further study is required to identify which are the most appropriate.

\subsection{Explanation for the TTEF Start Times Propagation}
\label{ssec:ttef}

Propagation on the lower and upper bounds of the start time variables $S_i$ are symmetric; Consequently we only present the case for the lower bounds' propagation.
To prune the lower bound of an activity~$u$, \ttef bounds propagation tentatively starts the activity~$u$ at its earliest start time~$\est{u}$ and then checks whether that causes a resource overload in any time window $[\est{a}, \lct{b})$ ($\{a, b\} \subseteq \setEF$).
Thus, bounds propagation and its explanation are very similar to that of the  consistency check.

The work of~\cite{Vilim:11} considers four positions of~$u$ relative to the time window: \emph{right} ($\est{a} \leq \est{u} < \lct{b} < \ect{u}$), \emph{inside} ($\est{a} \leq \est{u} < \ect{u} \leq \lct{b}$), \emph{through} ($\est{u} < \est{a} \wedge \lct{b} < \ect{u}$), and \emph{left} ($\est{u} < \est{a} < \ect{u} \leq \lct{b}$). The first two of these positions correspond to edge-finding propagation and the last two to extended edge-finding propagation.
We first consider only the right and inside positions, \ie $\est{a} \leq \est{u}$.
Note that $a$ could be $u$.
Then, 
\begin{equation}
R \cdot (\lct{b} - \est{a}) - energy(a,b,u) < 0 
\to \left\lceil \frac{rest(a,b,u)}{r_u} \right\rceil \leq S_u
\end{equation}
where $energy(a,b,u) := energy(a,b) - \energy{u}(a,b) + r_u \cdot (\min(\lct{b}, \ect{u}) - \est{u})$ and
\begin{multline*}
rest(a,b,u) := energy(a,b,u) - (R - r_u) \cdot (\lct{b} - \est{a}) \\
- r_u \cdot (\min(\lct{b}, \ect{u}) - \est{u})\enspace.
\end{multline*}
The first two terms in the sum of $energy(a,b,u)$ gives the energy consumption of all considered activities except~$u$, whereas the last term is the required energy of~$u$ if it is scheduled at~$\est{u}$ in the time window~$[\est{a}, \lct{b})$.
The propagation, including explanation generation, can be performed in $\OO(l^2 + k\cdot n)$ runtime, where $l = |\setEF|$ and $k$ the number of bounds' updates, if the resource profile is given. 
Moreover, \ttef propation does not necessarily consider each $u\in\setEF$, but those only that maximise $\min(\eEF{u}, r_u\cdot(\lct{b} - \est{a})) - r_u\cdot \max(0, \lct{b} - \lstEF{u})$ and satisfy $\est{a} \leq \est{u}$.
\begin{WithAppendix}
The corresponding algorithm is shown in Alg.~\ref{alg:ttef_filt} in App.~\ref{app:sec:ttef:alg}.
\end{WithAppendix}

A na\"{\i}ve explanation for a lower bound update from $\est{u}$ to $newLB := \lceil rest(a,b,u) / r_u \rceil$ with respect to the time window~$[\est{a}, \lct{b})$ additionally includes the previous and new lower bound on the left and right hand side of the implication, respectively, in comparison to the na\"{\i}ve explanation for a resource overload.
\begin{equation*}
\lit{\est{u} \leq S_u} \wedge \bigwedge_{i \in \setV \setminus \{u\} : p_i(a, b) > 0} 
	\lit{\est{i} \leq S_i} \wedge \lit{S_i \leq \lst{i}} 
	\to \lit{newLB \leq S_u}
\end{equation*}
As we discussed in the case of resource overload, we perform a similar generalisation for the activities in $\setV \setminus \{u\}$, and for~$u$ we decrease the lower bound on the left hand side as much as possible so that the same propagation holds when $u$ is executed at that decreased lower bound.
\begin{multline}
\lit{\est{a} + \lct{b} - newLB + 1 - p_u \leq S_u} \wedge \\ 
\bigwedge_{i \in \setV \setminus \{u\} : p_i(a, b) > 0} 
	\lit{\est{a} + p_i(a,b) - p_i \leq S_i} \wedge \lit{S_i \leq \lct{b} - p_i(a,b)} \\
	\to \lit{newLB \leq S_u}
\end{multline}
Again this more general explanation expresses the energetic reasoning propagation and the bounds of activities in $\{ i \in \setV \setminus \{u\} \mid p_i(a, b) > 0 \}$ can further be generalised in the same way as for a resource overload.
But here the available energy units $\Delta$ for widening the bounds is $rest(a,b,u) - r_u \cdot (newLB - 1) + 1$.
Hence, $0 \leq \Delta < r_u$ indicate that the explanation only can further be generalised a little bit.
We perform this generalisation as for the overload case.

\section{Experiments on Resource-constrained Project Scheduling Problems}

We carried out extensive experiments on \rcpsp instances
comparing our solution approach using both time-table and/or \ttef propagation.
We compare the obtained results on the lower bounds of the makespan with the best known so far.
Detailed results are available at \url{http://www.cs.mu.oz.au/~pjs/rcpsp}.

\begin{table}[tp] \small
\caption{Specifications of the benchmark suites.}
\label{tab:bench}
\centering
\begin{tabular}{ll|ccccp{60pt}}
suite & sub-suites & \#inst & \#act & $p_i$ & \#res & notes\\
\hline
\benchAT\cite{Alvarez:T:89} 
& \textsc{st}27/\textsc{st}51/\textsc{st}103 & 48 each & 25/49/101 & 1--12 & 6 each & \\
\hline
PSPLib~\cite{PSPLib:97}
& \benchJ{30}~\cite{Kolisch:SD:95}/\benchJ{60}/\benchJ{90} & 480 each & 30/60/90 & 1--10 & 4 each & \\
& \benchJ{120} & 600 & 30 & 1--10 & 4 & \\
\hline
\benchBL~\cite{Baptiste:LP:00}
& \textsc{bl}20/\textsc{bl}25 & 20 each & 20/25 & 1--6 & 3 each &\\
\hline
\benchPack~\cite{Carlier:N:03}
& & 55 & 15--33 & 1--19 & 2--5 &\\
\hline
\benchKSDD{15}~\cite{Kone:ALM:11}
& & 480 & 15 & 1--250 & 4 & based on \benchJ{30}\\
\hline
\benchPackD~\cite{Kone:ALM:11}
& & 55 & 15--33 & 1--1138 & 2--5 & based on \benchPack\\
\end{tabular}
\end{table}

We used six benchmark suites for which an overview is given in 
Tab.~\ref{tab:bench} where \#inst, \#act, $p_i$, and \#res are the number of
instances, number of activities, range of processing times, and number of
resources, respectively.
The first two suites are highly disjunctive, while the remainder are highly cumulative.

The experiments were run on a X86-64 architecture running GNU/Linux and a
Intel(R) Core(TM) i7 CPU processor at 2.8GHz.
The code was written in Mercury~\cite{Somogyi:HC:96} using the G12 Constraint
Programming Platform~\cite{G12:05}.

We model an instance as in~\cite{Schutt:FSW:11} using 
global cumulative constraints \cumu and difference logic constraints ($S_i + p_i \leq S_j$), resp.
In addition, between two activities $i$, $j$ in disjunction, 
\ie two activities which cannot concurrently run without overloading some resource, 
the two half-reified constraints~\cite{Feydy:SS:11} $b \to S_i + p_i \leq S_j$ and $\neg b \to S_j + p_j \leq S_i$ are posted where $b$ is a Boolean variable.

We run cumulative constraint propagation using different phases:
\begin{itemize}
\item[(a)] time-table consistency check in $\OO(n + p \log p)$ runtime,
\item[(b)] \ttef consistency check in $\OO(l^2 + n)$ runtime as defined in Section~\ref{ssec:ttefc},
\item[(c)] time-table bounds' propagation in $\OO(l\cdot p + k \cdot
  \min(R,n))$ runtime,
 and 
\item[(d)] \ttef bounds' propagation in $\OO(l^2 + k\cdot n)$ runtime as
  defined in Section~\ref{ssec:ttef} where $k, l, n, p$ are the numbers of
  bounds' updates, unfixed activities, all activities, and height changes in
  the resource profile, resp.
\end{itemize}
Note that in our setup phase (d) \ttef bounds' propagation does not take
into account the bounds' changes of the phase (c) time-table bounds'
propagation.
For the experiments, we consider three settings of the cumulative propagator:
\ttProp executes phases (a) and (c), \ttefCheck (a--c), and \ttefProp
(a--d).  Note that phases (c)  and (d) are not run if either phase (a) or
(b) detects inconsistency.

\subsection{Upper Bound Computation}

For solving \rcpsp we use the same branch-and-bound algorithm as we used
in~\cite{Schutt:FSW:11}, 
but here we limit ourselves to the search heuristic 
\hr which was the most robust one in our previous studies~\cite{Schutt:FSW:09,Schutt:FSW:11}.
It executes an adapted search of~\cite{Baptiste:LP:00} using serial 
scheduling generation for the first 500 choice points and, then, continues 
with an activity based search (a variant of \vsids~\cite{Moskewicz:MZZM:01})
on the Boolean variables representing a lower part $x \leq v$ and upper part $v < x$ of the variable $x$'s domain where $x$ is either a start time or the makespan variable and $v$ a value of $x$'s initial domain.
Moreover, it is interleaved with a geometric restart policy~\cite{Walsh:99} 
on the number of node failures for which the restart base and factor are 250 
failures and 2.0, respectively.
The search was halted after 10 minutes.

The results are given in Tab.~\ref{tab:ub:disj} and~\ref{tab:ub:cumu}. For each
benchmark suite, the number of solved instances (\#svd) is given. The column
cmpr($a$) shows the results on the instances solved by all methods, where $a$ is
the number of such instances. 
The left entry in that column is the average runtime on these instances in
seconds, and the right entry is the average number of failures during search.
The entries in column all($a$) have the same meaning, 
but here all instances are considered where $a$ is the total number of
instances. For unsolved instances, the number of failures after 10 minutes is
used.

\begin{table}[tp] \small
\centering

\caption{UB results on highly disjunctive \rcpsp{}s.}
\label{tab:ub:disj}
\begin{tabular}{l||c|rr|rr||c|rr|rr}
& \multicolumn{5}{c||}{\benchJ{30}} & \multicolumn{5}{c}{\benchJ{60}}\\
& \#svd & \multicolumn{2}{c|}{cmpr(480)} & \multicolumn{2}{c||}{all(480)} 
& \#svd & \multicolumn{2}{c|}{cmpr(429)} & \multicolumn{2}{c}{all(480)}\\ 
\hline
\ttProp 
&  480 & 0.12 & 1074 & 0.12 & 1074
&  430 & 1.82 & 5798 & 64.25 & 93164\\
\ttefCheck 
&  480 & 0.20 & 1103 & 0.20 & 1103
&  431 & 2.00 & 4860 & 64.39 & 80845\\
\ttefProp 
&  480 & 0.23 & 991 & 0.23 & 991
&  432 & 3.04 & 5191 & 64.87 & 62534\\
\hline \hline
& \multicolumn{5}{c||}{\benchJ{90}} & \multicolumn{5}{c}{\benchJ{120}}\\
& \#svd & \multicolumn{2}{c|}{cmpr(400)} & \multicolumn{2}{c||}{all(480)}
& \#svd & \multicolumn{2}{c|}{cmpr(280)} & \multicolumn{2}{c}{all(600)}\\
\hline
\ttProp 
&  400 & 5.03 & 9229 & 104.09 & 132234
&  283 & 9.71 & 15022 & 322.35 & 398941\\
\ttefCheck 
&  400 & 6.93 & 9512 & 105.69 & 104297
&  282 & 13.47 & 16958 & 324.73 & 297562\\
\ttefProp 
&  400 & 8.10 & 8830 & 106.66 & 72402
&  283 & 14.97 & 13490 & 324.66 & 186597\\
\hline \hline
& \multicolumn{5}{c||}{\benchAT}\\
& \#svd & \multicolumn{2}{c|}{cmpr(129)} & \multicolumn{2}{c||}{all(144)}\\
\cline{1-6}
\ttProp 
&  132 & 8.90 & 19997 & 66.22 & 87226\\
\ttefCheck 
&  130  & 9.36 & 16466 & 69.41 & 72056\\
\ttefProp 
&  129  & 13.55 & 17239 & 74.60 & 63554
\end{tabular}
\bigskip

\caption{UB results on highly cumulative \rcpsp{}s.}
\label{tab:ub:cumu}
\begin{tabular}{l||c|rr|rr||c|rr|rr}
& \multicolumn{5}{c||}{\benchBL} & \multicolumn{5}{c}{\benchPack}\\
& \#svd & \multicolumn{2}{c|}{cmpr(40)} & \multicolumn{2}{c||}{all(40)} 
& \#svd & \multicolumn{2}{c|}{cmpr(16)} & \multicolumn{2}{c}{all(55)}\\ 
\hline
\ttProp 
&  40 & 0.16 & 2568 & 0.16 & 2568
&  16 & 77.65 & 245441 & 447.69 & 699615\\
\ttefCheck 
&  40 & 0.02 & 370 & 0.02 & 370
&  39 & 37.22 & 122038 & 186.79 & 292101\\
\ttefProp 
&  40 & 0.02 & 269 & 0.02 & 269
&  39 & 44.44 & 105751 & 188.23 & 257747\\
\hline \hline
& \multicolumn{5}{c||}{\benchKSDD{15}} & \multicolumn{5}{c}{\benchPackD}\\
& \#svd & \multicolumn{2}{c|}{cmpr(480)} & \multicolumn{2}{c||}{all(480)}
& \#svd & \multicolumn{2}{c|}{cmpr(37)} & \multicolumn{2}{c}{all(55)}\\ 
\hline
\ttProp 
&  480 & 0.01 & 26 & 0.01 & 26
&  37 & 32.72 & 42503 & 218.26 & 184293\\
\ttefCheck 
&  480 & 0.01 & 26 & 0.01 & 26
&  37 & 23.96 & 32916 & 212.37 & 170301\\
\ttefProp 
&  480 & 0.01 & 26 & 0.01 & 26
&  37 & 36.93 & 37004 & 221.11 & 157015\\
\end{tabular}

\end{table}

Table~\ref{tab:ub:disj} shows the results on the highly disjunctive \rcpsp{}s. 
As expected, the stronger propagation (\ttefCheck, \ttefProp) reduces the search space overall in comparision to \ttProp, but the average runtime is higher by a factor of about 5\%--70\% and 50\%--100\% for \ttefCheck and \ttefProp.
Interestingly, \ttefCheck and \ttefProp solved respectively 1 and 2 more
instances on \benchJ{60} and closed the instance j120\_1\_1 on \benchJ{120} which has an optimal makespan 105. 
This makespan corresponds to the best known upper bound.
However, the stronger propagation does not generally
pay off for a \cp solver with nogood learning.

Table~\ref{tab:ub:cumu} presents the results on highly cumulative \rcpsp{}s which clearly shows the benefit of \ttef propagation, especially on \benchBL for which \ttefCheck and \ttefProp reduce the search space and the average runtime by a factor of 8, and \benchPack for which they solved 23 instances more than \ttProp.
On \benchPackD, \ttefCheck is about 50\% faster 
on average than \ttProp while \ttefProp is slightly slower on average than \ttProp.
No conclusion can be drawn on \benchKSDD{15} because the instances are easy for \lcg solvers.

\subsection{Lower Bound Computation}

The lower bound computation tries to solve \rcpsp{}s in a destructive way by
converging to the optimal \emph{makespan} from below, \ie it repeatedly proves that there exists no solution for current makespan considered and continues with an incremented \emph{makespan} by 1.
If a solution found then it is the optimal one.
For these experiments we use the search heuristic~\hs as we did in~\cite{Schutt:FSW:09,Schutt:FSW:11}.
This heuristic is \hr (as decribed earlier) but no restart.
We used the same parameters as for \hr.
For the starting \emph{makespan}, 
we choose the best known lower bounds on \benchJ{60}, \benchJ{90}, and \benchJ{120} recorded in the PSPLib at \url{http://129.187.106.231/psplib/} and~\cite{Vilim:11} at \url{http://vilim.eu/petr/cpaior2011-results.txt}. 
On the other suites, the search starts from \emph{makespan} 1.
Due to the tighter \emph{makespan}, it is expected that the \ttef
propagation will perform better than for upper bound computation on the highly disjunctive instances.
The search was cut off at 10 minutes as in~\cite{Schutt:FSW:09,Schutt:FSW:11}.
 
\begin{table}[tp] \small
\centering
\caption{LB results on \benchAT, \benchPack, and \benchPackD}
\label{tab:lb:others}
\ignore{
\begin{tabular}{l|lr|lr|lr}
& \multicolumn{2}{c|}{\benchAT} & \multicolumn{2}{c|}{\benchPack} & \multicolumn{2}{c}{\benchPackD} \\
\hline
\ttefCheck & 5/2/5 & +70 & 1/3/12 & +125 & 1/3/14 & +2324 \\
\ttefProp & 8/0/4 & +62 & 2/2/12 & +127 & 3/2/13 & +2202
\end{tabular}
}
\begin{tabular}{l|lr|lr|lr}
& \multicolumn{2}{c|}{\benchAT} & \multicolumn{2}{c|}{\benchPack} & \multicolumn{2}{c}{\benchPackD} \\
\hline
\ttefCheck & 5/4/3 & +52 & 0/4/12 & +100 & 0/7/11 & +632 \\
\ttefProp & 7/2/3 & +44 & 1/4/11 & +101 & 2/5/10 & +618
\end{tabular}
\bigskip

\caption{LB results on \benchJ{60}, \benchJ{90}, and \benchJ{120}}
\label{tab:lb:psplib}
\ignore{
\begin{tabular}{l|rr|rrrr|rrrrrrrr}
& \multicolumn{2}{c|}{\benchJ{60}} & \multicolumn{4}{c|}{\benchJ{90}} & \multicolumn{8}{c}{\benchJ{120}} \\
& +1 & +2 & +1 & +2 & +3 & +4 & +1 & +2 & +3 & +4 & +5 & +6 & +7 & +8 \\
\hline
\ttefCheck & 4 & 1  & 12 & 1  & - & -  & 28 & 8 & 4 & - & - & - & 2 & - \\
\ttefProp  & 7 & 5  & 25 & 14 & 3 & 1  & 91 & 20 & 10 & 5 & 2 & - & - & 2
\end{tabular}
}
\begin{tabular}{ll|rrr|rrrrr|rrrrrrrrrr}
& & \multicolumn{3}{c|}{\benchJ{60}} & \multicolumn{5}{c|}{\benchJ{90}} & \multicolumn{10}{c}{\benchJ{120}} \\
& & +1 & +2 & +3 & +1 & +2 & +3 & +4 & +5 & +1 & +2 & +3 & +4 & +5 & +6 & +7 & +8 & +9 & +10\\
\hline
\multirow{2}{*}{1 min}
& \ttefCheck & 4  & 1  & - & 12 &  1 & - & - & -  & 27 & 8  &  4 & - & - & - & 2 & - & - & -\\
& \ttefProp  & 7  & 5  & - & 25 & 14 & 3 & 1 & -  & 90 & 20 & 10 & 5 & 2 & - & - & 2 & - & -\\
\hline
\multirow{2}{*}{10 mins}
& \ttefCheck & 21 & 2 & -  & 25 &  7 & - & - & -  & 68  & 16 & 4 & 4 & 2 & - & - & 1 & 1 & -\\
& \ttefProp  & 13 & 6 & 3  & 35 & 17 & 6 & 3 & 1  & 116 & 39 & 9 & 9 & 4 & 1 & - & - & 1 & 1
\end{tabular}
\end{table}

Table~\ref{tab:lb:others} 
shows the results on \benchAT, \benchPack, and \benchPackD restricted to the
instances that none of the methods could solve using the upper bound computation, 
that are 12, 16, and 18 for \benchAT, \benchPack, and \benchPackD, respectively.
An entry $a$/$b$/$c$ for method~$x$ means that $x$ achieved respectively 
$a$-times, $b$-times and $c$-times a worse, the same and a better lower bound
than \ttProp. The entry +$d$ is the sum of lower bounds' differences of method $x$ to \ttProp.
On \benchPack and \benchPackD, \ttefCheck and \ttefProp clearly perform better than \ttProp.
On the highly disjunctive instances in \benchAT, \ttefCheck and \ttProp are almost
balanced whereas \ttProp could generate better lower bounds on more instances as \ttefProp.
The lower bounds' differences on \benchAT are dominated 
by the instance \texttt{st103\_4} 
for which \ttefCheck and \ttefProp retrieved a lower bound improvement of 54 and 53 time periods with respect to \ttProp.

The more interesting results are presented in Tab.~\ref{tab:lb:psplib}
because 
the best lower bounds are known for all the remaining open instances (48, 77, 307 in
\benchJ{60}, \benchJ{90}, \benchJ{120}).\footnote{Note that the PSPLib still
  lists the instances \texttt{j60\_25\_5}, \texttt{j90\_26\_5}, \texttt{j120\_8\_3}, \texttt{j120\_48\_5}, and
  \texttt{j120\_35\_5} as open, but we closed the first four ones
  in~\cite{Schutt:FSW:11} and \cite{Liess:M:08} closed the last one.} 
An entry in a column +$d$ shows the number of instances for that the corresponding method could improve the lower bound by $d$ time periods.
On these instances, we run at first the experiments with a runtime limit of one minute as it was done in the experiments for \ttef propagation in~\cite{Vilim:11} but he used a \cp solver without nogood learning. 
\ttProp could not improve any lower bound because its corresponding results are already recorded in the PSPLib.
\ttefCheck and \ttefProp 
improved the lower bounds of 59 and 183 instances, respectively, 
which is about 13.7\% and 42.4\% of the open instances.
Although, the experiments in \cite{Vilim:11} were run on a slower machine\footnote{Intel(R) Core(TM)2 Duo CPU T9400 on 2.53GHz} the results confirm 
the importance of nogood learning.
For the experiments with 10 minutes runtime, we excluded \ttProp due to time constraints and expected inferior results to \ttefCheck and \ttefProp.
With the extended runtime, \ttefCheck and \ttefProp could improved 
the lower bounds of more instance, namely 151 and 264 instances, respectively, 
which is about 35.0\% and 61.1\%.
Moreover, 3, 1, and 1 of the remaining open instances on \benchJ{60}, \benchJ{90}, and \benchJ{120}, respectively, could be solved optimally.
\begin{WithAppendix}
See App.~\ref{app:sec:newlb} for the listing of the closed instances and the new lower bounds.
\end{WithAppendix}

\section{Conclusion and Outlook}

We present explanations for the recently developed \ttef propagation of the global cumulative constraint for lazy clause generation solvers.
These explanations express an energetic reasoning propagation which is a 
stronger propagation than the \ttef one.

Our implementation of this propagator was compared to 
time-table propagation in lazy clause generation solvers on six benchmark suites.
The preliminary results confirms the importance of energy-based reasoning on highly disjunctive \rcpsp{}s for \cp solvers with nogood learning.

Moreover, our approach with \ttef propagation was able to close one instance. It
also improves the best known lower bounds for 264 of the remaining 432 remaining
open instances on \rcpsp{}s from the PSPLib.

In the future, we want to integrate the extended edge-finding 
propagation into \ttef propagation as it was originally proposed
in~\cite{Vilim:11}, to perform experiments on cutting and packing problems, and
to study different variations of explanations for \ttef propagation.
Furthermore, we want to look at a more efficient implementation of the \ttef propagation as well as an implementation of energetic reasoning.

\paragraph{Acknowledgements}

NICTA is funded by the Australian Government as represented by the Department of Broadband, Communications and the Digital Economy and the Australian Research Council through the ICT Centre of Excellence program.
This work was partially supported by Asian Office of Aerospace Research and Development grant 10-4123.

\bibliographystyle{plainnat}
\bibliography{ttef_expl_refs}

\clearpage
\begin{WithAppendix}
\appendix

\section{TTEF propagation algorithms}
\label{app:sec:ttef:alg}

Algorithm~\ref{alg:ttef_check} shows the \ttef consistency check used.
The outer loop (lines 2--16) iterates over all distinctive possible end times for the time windows while the inner loop (lines 7--16) iterates over all possible start times.
In line 11 (12), it checks whether $a$ must fully (partially) be executed in
the current time window and further ones checked in the same inner loop. If
so it adds the required free energy units $\eEF{a}$ of $a$ to $E$.
In line 13, it calculates the still available energy units in the time
window $[begin, end)$ 
taking the energy units from the resource profile $\ttEn(a,b)$ into account.
If this results in a resource overload then a corresponding explanation is generated (line 15) and the algorithm fails; otherwise, the algorithm succeeds.

\begin{algorithm}[thbp]
	\caption{\ttef consistency check.}
	\label{alg:ttef_check}
	\SetKwInOut{Input}{Input}
	\SetKw{KwDownTo}{down to}
	\SetKw{KwAnd}{and}
	\SetKw{KwOr}{or}
	\footnotesize
	\Input{$X$ an array of activities sorted in non-decreasing order of the earliest start time.}
	\Input{$Y$ an array of activities sorted in non-decreasing order of the latest completion time.}
	$end = \infty$\;
	\For{$y := n$ \KwDownTo $1$}{
		$b := Y[y]$\;
		\lIf{$\lct{b} = end$}{continue}\;
		$end := \lct{b}$\;
		$E := 0$\;
		\For{$x := n$ \KwDownTo $1$}{
			$a := X[x]$\;
			\lIf{$end \leq \est{a}$}{continue}\;
			$begin := \est{a}$\;
			\lIf{$\lct{a} \leq end$}{
				$E := E + \eEF{a}$\;
			}
			\lIf{$\lstEF{a} < end$}{			$E := E + r_a \cdot (end - \lstEF{a})$}\;
			$avail := R \cdot (end - begin) - E - \ttEn(a,b)$\;
			\If{$avail < 0$}{
				explainOverload($begin$, $end$)\;
				\Return{false}\;
			}
		}
	}
	\Return{true}\;
\end{algorithm}

Algorithm~\ref{alg:ttef_filt} shows the lower bounds propagation algorithm.
As for Alg.~\ref{alg:ttef_check} the outer loop (lines 3--24) and inner loop
(lines 7--24) iterate 
over the end and start times of the time windows $[begin, end)$, but require more book keeping.
In line 6, it initialises $E$. $u$, and $enReqU$ where: $E$ records the
required energy units by the considered activities that must fully or
partially be run in the time window; and $u$ stores the activity that maximises
$\min(\eEF{u}, r_u\cdot (end - begin)) - r_u\cdot \max(0, end - \lstEF{u})$ and that value is saved in $enReqU$.
If $a$ must be fully or partially be executed 
in the time window then the corresponding energy units are added to $E$ in lines 11 and 14, resp.
The desired activity for pruning is computed in lines 13, 15, and 16, 
whereas the available energy units are calculated in line 17.
In the case that there is not sufficient energy available then the condition
of line 18 holds and the algorithm determines the first possible start time for~$u$ (lines 19, 20).
If that is larger than the recorded earliest start time in $\est{u}'$ then the algorithm generates the explanation (line 22) and postpones the update (line 23) after finishing with the outer loop (line 25).
\begin{algorithm}[tbp]
	\caption{\ttef lower bounds propagator on the start times.}
	\label{alg:ttef_filt}
	\SetKwInOut{Input}{Input}
	\SetKw{KwDownTo}{down to}
	\SetKw{KwAnd}{and}
	\SetKw{KwOr}{or}
	\footnotesize
	\Input{$X$ an array of activities sorted in non-decreasing order of the earliest start time.}
	\Input{$Y$ an array of activities sorted in non-decreasing order of the latest completion time.}
	\lFor{$i \in \setEF$}{$\est{i}' := \est{i}$}\;
	$end := \infty$; $k := 0$\;
	\For{$y := n$ \KwDownTo $1$}{
		$b := Y[y]$\;
		\lIf{$\lct{b} = end$}{continue}\;
		$end := \lct{b}$;
		$E := 0$;
		$u := -\infty$; 
		$enReqU := 0$\;
		\For{$x := n$ \KwDownTo $1$}{
			$a := X[x]$\;
			\lIf{$end \leq \est{a}$}{continue}\;
			$begin := \est{a}$\;
			\lIf{$\lct{a} \leq end$}{
				$E := E + \eEF{a}$\;
			}
			\Else{
				$enIn := r_a \cdot \max(0, end - \lstEF{a})$\;
				$E := E + enIn$\;
				$enReqA := \min(\eEF{a}, r_a \cdot (end - \est{a})) - enIn$\;
				\lIf{$enReqA > enReqU$}{$u := a$; $enReqU := enReqA$}\;
			}
			$avail := R \cdot (end - begin) - E - \ttEn(a, b)$\;
			\If{$enReqU > 0$ \KwAnd $avail - enReqU < 0$}{
				$rest := E - avail - r_a \cdot \max(0, end - \lst{a})$\;
				$lbU := begin + \lceil rest / r_u \rceil$\;
				\If{$\est{u}' < lbU$}{
					$expl := explainUpdate(begin, end, u, \est{u}', lbU)$\;
					$Update[$++$k] := (u, lbU, expl)$\;
					$\est{u}' := lbU$\;
				}
			}
		}
	}
	\lFor{$z := 1$ \KwTo $k$}{$updateLB(Update[z])$}\;
\end{algorithm}

\section{Closed Instances and New Lower Bounds on PSPLib}
\label{app:sec:newlb}

From the open instances, we closed the instances 9\_3 (100), 9\_9 (99), 25\_10 (108) on \benchJ{60}, 5\_6 (86) on \benchJ{90}, and 1\_1 (105), 8\_6 (85) on \benchJ{120} where the number in brackets shows the optimal makespan.
We computed new lower bounds on the remaining open instances from the PSPLib.
Tables~\ref{tab:newlb:j60}--\ref{tab:newlb:j120} list these new lower bounds where the column ``inst'' shows the name of the instance and the column ``LB'' the corresponding new lower bound.

\ignore{
\begin{table} \small
\centering
\caption{New lower bounds on \benchJ{60}.}
\label{tab:newlb:j60}
\begin{tabular}{cc|cc|cc|cc|cc|cc|cc}
inst & LB & inst & LB & inst & LB & inst & LB & inst & LB & inst & LB & inst & LB\\
\hline
9\_5 & 81  & 9\_7 & 103  & 9\_10 & 89  & 13\_3 & 84  & 13\_4 & 98  & 13\_7 & 82  & 13\_9 & 96\\
13\_10 & 113  & 29\_1 & 97  & 29\_7 & 115  & 29\_8 & 97  & 41\_10 & 106
\end{tabular}
\end{table}
}

\begin{table} \small
\centering
\caption{New lower bounds on \benchJ{60}.}
\label{tab:newlb:j60}
\begin{tabular}{cc|cc|cc|cc|cc|cc|cc}
inst & LB & inst & LB & inst & LB & inst & LB & inst & LB & inst & LB & inst & LB\\
\hline
9\_1 & 85  & 9\_5 & 81  & 9\_6 & 106  & 9\_7 & 103  & 9\_8 & 95  & 9\_10 & 89  & 13\_1 & 105\\
13\_2 & 103  & 13\_3 & 84  & 13\_4 & 98  & 13\_7 & 82  & 13\_8 & 115  & 13\_9 & 96  & 13\_10 & 113\\
25\_2 & 96  & 25\_4 & 106  & 25\_6 & 106  & 29\_1 & 97  & 29\_6 & 144  & 29\_7 & 115  & 29\_8 & 97\\
41\_3 & 90  & 41\_10 & 106  & 45\_1 & 90
\end{tabular}
\end{table}

\ignore{
\begin{table} \small
\centering
\caption{New lower bounds on \benchJ{90}.}
\label{tab:newlb:j90}
\begin{tabular}{cc|cc|cc|cc|cc|cc|cc}
inst & LB & inst & LB & inst & LB & inst & LB & inst & LB & inst & LB & inst & LB\\
\hline
5\_3 & 84  & 5\_5 & 109  & 5\_7 & 106  & 5\_8 & 97  & 9\_3 & 98  & 9\_4 & 120  & 9\_5 & 127\\
9\_6 & 113  & 9\_7 & 103  & 9\_8 & 111  & 9\_9 & 106  & 9\_10 & 105  & 13\_2 & 119  & 13\_8 & 113\\
13\_9 & 117  & 25\_1 & 117  & 25\_2 & 122  & 25\_3 & 113  & 25\_4 & 128  & 25\_6 & 113  & 25\_8 & 131\\
25\_9 & 98  & 25\_10 & 119  & 29\_1 & 126  & 29\_2 & 122  & 29\_4 & 139  & 29\_6 & 117  & 29\_7 & 160\\
29\_8 & 146  & 41\_1 & 129  & 41\_2 & 154  & 41\_3 & 149  & 41\_4 & 142  & 41\_5 & 116  & 41\_6 & 124\\
41\_7 & 145  & 41\_8 & 148  & 41\_9 & 110  & 41\_10 & 144  & 45\_1 & 143  & 45\_2 & 138  & 45\_6 & 163\\
46\_9 & 86
\end{tabular}
\end{table}
}

\begin{table} \small
\centering
\caption{New lower bounds on \benchJ{90}.}
\label{tab:newlb:j90}
\begin{tabular}{cc|cc|cc|cc|cc|cc|cc}
inst & LB & inst & LB & inst & LB & inst & LB & inst & LB & inst & LB & inst & LB\\
\hline
5\_3 & 84  & 5\_5 & 109  & 5\_7 & 106  & 5\_8 & 97  & 5\_9 & 114  & 5\_10 & 95  & 9\_2 & 122\\
9\_3 & 98  & 9\_4 & 120  & 9\_5 & 127  & 9\_6 & 113  & 9\_7 & 103  & 9\_8 & 111  & 9\_9 & 106\\
9\_10 & 105  & 13\_2 & 119  & 13\_3 & 105  & 13\_5 & 109  & 13\_7 & 116  & 13\_8 & 113  & 13\_9 & 117\\
13\_10 & 114  & 21\_7 & 106  & 21\_8 & 108  & 25\_1 & 117  & 25\_2 & 122  & 25\_3 & 113  & 25\_4 & 128\\
25\_5 & 110  & 25\_6 & 113  & 25\_8 & 131  & 25\_9 & 98  & 25\_10 & 119  & 29\_1 & 126  & 29\_2 & 122\\
29\_4 & 139  & 29\_6 & 117  & 29\_7 & 160  & 29\_8 & 146  & 29\_9 & 120  & 30\_9 & 92  & 37\_2 & 114\\
41\_1 & 129  & 41\_2 & 154  & 41\_3 & 149  & 41\_4 & 142  & 41\_5 & 116  & 41\_6 & 124  & 41\_7 & 145\\
41\_8 & 148  & 41\_9 & 110  & 41\_10 & 144  & 45\_1 & 143  & 45\_2 & 138  & 45\_3 & 144  & 45\_4 & 126\\
45\_6 & 163  & 45\_7 & 129  & 45\_8 & 150  & 45\_9 & 145  & 45\_10 & 156  & 46\_9 & 86
\end{tabular}
\end{table}

\begin{table} \small
\centering
\caption{New lower bounds on \benchJ{120}.}
\label{tab:newlb:j120}
\begin{tabular}{cc|cc|cc|cc|cc|cc|cc}
inst & LB & inst & LB & inst & LB & inst & LB & inst & LB & inst & LB & inst & LB\\
\hline
6\_1 & 134  & 6\_2 & 127  & 6\_5 & 117  & 6\_6 & 141  & 6\_8 & 141  & 6\_9 & 150  & 6\_10 & 158\\
7\_1 & 99  & 7\_3 & 98  & 7\_4 & 106  & 7\_6 & 116  & 7\_7 & 114  & 7\_8 & 93  & 7\_9 & 87\\
7\_10 & 112  & 8\_2 & 102  & 8\_5 & 100  & 8\_9 & 90  & 8\_10 & 92  & 9\_4 & 85  & 11\_1 & 157\\
11\_2 & 147  & 11\_3 & 189  & 11\_4 & 178  & 11\_5 & 194  & 11\_6 & 192  & 11\_7 & 149  & 11\_8 & 153\\
11\_10 & 164  & 12\_1 & 126  & 12\_2 & 112  & 12\_4 & 122  & 12\_5 & 155  & 12\_6 & 116  & 13\_1 & 124\\
13\_3 & 116  & 13\_4 & 109  & 13\_6 & 96  & 13\_9 & 83  & 14\_2 & 91  & 14\_5 & 94  & 14\_7 & 90\\
16\_1 & 181  & 16\_3 & 221  & 16\_4 & 191  & 16\_6 & 195  & 16\_8 & 183  & 17\_5 & 124  & 17\_6 & 134\\
18\_8 & 102  & 18\_9 & 89  & 18\_10 & 97  & 26\_1 & 155  & 26\_2 & 159  & 26\_3 & 158  & 26\_4 & 161\\
26\_5 & 139  & 26\_6 & 171  & 26\_7 & 147  & 26\_8 & 168  & 26\_9 & 161  & 26\_10 & 178  & 27\_1 & 107\\
27\_2 & 110  & 27\_3 & 142  & 27\_4 & 105  & 27\_5 & 106  & 27\_6 & 133  & 27\_7 & 119  & 27\_8 & 136\\
27\_9 & 121  & 27\_10 & 111  & 28\_1 & 106  & 31\_1 & 181  & 31\_2 & 176  & 31\_3 & 160  & 31\_4 & 195\\
31\_5 & 187  & 31\_6 & 182  & 31\_7 & 191  & 31\_8 & 176  & 31\_9 & 176  & 31\_10 & 202  & 32\_1 & 144\\
32\_2 & 123  & 32\_5 & 133  & 32\_6 & 122  & 32\_8 & 132  & 33\_1 & 105  & 33\_2 & 107  & 33\_3 & 102\\
33\_4 & 107  & 33\_8 & 107  & 33\_9 & 109  & 34\_1 & 76  & 34\_2 & 103  & 34\_3 & 99  & 34\_5 & 102\\
36\_1 & 201  & 36\_3 & 218  & 36\_5 & 213  & 36\_7 & 196  & 36\_9 & 203  & 37\_2 & 141  & 37\_5 & 195\\
37\_8 & 169  & 37\_9 & 138  & 38\_1 & 105  & 38\_2 & 119  & 38\_4 & 138  & 38\_6 & 119  & 38\_7 & 103\\
38\_10 & 137  & 39\_2 & 105  & 40\_1 & 80  & 42\_1 & 107  & 46\_1 & 172  & 46\_2 & 187  & 46\_3 & 163\\
46\_5 & 136  & 46\_7 & 158  & 46\_9 & 157  & 46\_10 & 175  & 47\_1 & 130  & 47\_3 & 119  & 47\_4 & 120\\
47\_5 & 126  & 47\_6 & 128  & 47\_7 & 114  & 47\_8 & 124  & 47\_10 & 128  & 48\_4 & 123  & 51\_1 & 186\\
51\_2 & 200  & 51\_3 & 193  & 51\_4 & 197  & 51\_6 & 193  & 51\_7 & 185  & 51\_8 & 186  & 51\_9 & 190\\
51\_10 & 201  & 52\_1 & 161  & 52\_2 & 169  & 52\_3 & 126  & 52\_4 & 157  & 52\_5 & 158  & 52\_6 & 183\\
52\_7 & 142  & 52\_8 & 148  & 52\_9 & 142  & 52\_10 & 131  & 53\_1 & 138  & 53\_2 & 109  & 53\_4 & 138\\
53\_5 & 109  & 53\_6 & 101  & 53\_8 & 135  & 53\_10 & 124  & 54\_1 & 102  & 54\_5 & 107  & 54\_6 & 104\\
54\_8 & 100  & 54\_9 & 105  & 57\_1 & 173  & 57\_2 & 151  & 57\_3 & 176  & 57\_5 & 170  & 57\_6 & 176\\
57\_7 & 156  & 57\_9 & 157  & 58\_2 & 122  & 58\_3 & 117  & 58\_4 & 138  & 58\_5 & 116  & 58\_6 & 135\\
58\_7 & 143  & 58\_8 & 126  & 58\_9 & 126  & 59\_5 & 104  & 59\_6 & 112  & 59\_8 & 107  & 59\_9 & 117\\
59\_10 & 128  & 60\_3 & 88  & 60\_7 & 91
\end{tabular}
\end{table}

\section{Best Lower and Upper Bounds Retrieved}

For a later comparison, Tables~\ref{tab:lub:at}--\ref{tab:lub:packd} show 
the best lower and upper bounds for \benchAT, \benchPack, and \benchPackD retrieved by 
one of the methods \ttProp, \ttefCheck, and \ttefProp.
The column ``inst'' shows the instance name and the column ``LB/UB'' the 
corresponding lower and upper bound. If these bounds are equal then only
one number is given.

\begin{table} \small
\centering
\caption{Lower and upper bounds for \benchAT.}
\label{tab:lub:at}
\begin{tabular}{cc|cc|cc|cc|cc}
inst & LB/UB & inst & LB/UB & inst & LB/UB & inst & LB/UB & inst & LB/UB\\
\hline
27\_1 & 41 & 27\_2 & 53 & 27\_3 & 68 & 27\_4 & 112/114 & 27\_5 & 56\\
27\_6 & 73 & 27\_7 & 54 & 27\_8 & 95 & 27\_9 & 38 & 27\_10 & 45\\
27\_11 & 57 & 27\_12 & 73 & 27\_13 & 38 & 27\_14 & 55 & 27\_15 & 46\\
27\_16 & 75 & 27\_17 & 55 & 27\_18 & 55 & 27\_19 & 79 & 27\_20 & 152\\
27\_21 & 92 & 27\_22 & 86 & 27\_23 & 82 & 27\_24 & 106 & 27\_25 & 51\\
27\_26 & 53 & 27\_27 & 58 & 27\_28 & 95 & 27\_29 & 51 & 27\_30 & 76\\
27\_31 & 75 & 27\_32 & 82 & 27\_33 & 66 & 27\_34 & 61 & 27\_35 & 115\\
27\_36 & 146 & 27\_37 & 78 & 27\_38 & 100 & 27\_39 & 119 & 27\_40 & 130\\
27\_41 & 60 & 27\_42 & 53 & 27\_43 & 75 & 27\_44 & 88 & 27\_45 & 49\\
27\_46 & 65 & 27\_47 & 75 & 27\_48 & 80 & 51\_1 & 98 & 51\_2 & 96\\
51\_3 & 133 & 51\_4 & 161/219 & 51\_5 & 97 & 51\_6 & 126 & 51\_7 & 120\\
51\_8 & 194 & 51\_9 & 74 & 51\_10 & 73 & 51\_11 & 99 & 51\_12 & 116/137\\
51\_13 & 84 & 51\_14 & 86 & 51\_15 & 86 & 51\_16 & 132 & 51\_17 & 84\\
51\_18 & 99 & 51\_19 & 170 & 51\_20 & 274 & 51\_21 & 145 & 51\_22 & 168\\
51\_23 & 183 & 51\_24 & 228 & 51\_25 & 95 & 51\_26 & 89 & 51\_27 & 113\\
51\_28 & 164 & 51\_29 & 98 & 51\_30 & 105 & 51\_31 & 130 & 51\_32 & 139\\
51\_33 & 116 & 51\_34 & 115 & 51\_35 & 173 & 51\_36 & 300 & 51\_37 & 162\\
51\_38 & 177 & 51\_39 & 189 & 51\_40 & 218 & 51\_41 & 102 & 51\_42 & 108\\
51\_43 & 121 & 51\_44 & 174 & 51\_45 & 122 & 51\_46 & 125 & 51\_47 & 151\\
51\_48 & 167 & 103\_1 & 158 & 103\_2 & 182 & 103\_3 & 216/259 & 103\_4 & 280/445\\
103\_5 & 191 & 103\_6 & 207/209 & 103\_7 & 234/293 & 103\_8 & 207/294 & 103\_9 & 139\\
103\_10 & 119 & 103\_11 & 160/169 & 103\_12 & 213/302 & 103\_13 & 127 & 103\_14 & 152\\
103\_15 & 157/168 & 103\_16 & 167/179 & 103\_17 & 209 & 103\_18 & 232 & 103\_19 & 301\\
103\_20 & 475 & 103\_21 & 276 & 103\_22 & 295 & 103\_23 & 368 & 103\_24 & 449\\
103\_25 & 177 & 103\_26 & 183 & 103\_27 & 199 & 103\_28 & 295 & 103\_29 & 225\\
103\_30 & 231 & 103\_31 & 227 & 103\_32 & 281 & 103\_33 & 220 & 103\_34 & 264\\
103\_35 & 341 & 103\_36 & 575 & 103\_37 & 327 & 103\_38 & 376 & 103\_39 & 389\\
103\_40 & 451 & 103\_41 & 191 & 103\_42 & 187 & 103\_43 & 260 & 103\_44 & 375\\
103\_45 & 216 & 103\_46 & 251 & 103\_47 & 262 & 103\_48 & 300
\end{tabular}
\end{table}

\begin{table} \small
\centering
\caption{Lower and upper bounds for \benchPack.}
\label{tab:lub:pack}
\begin{tabular}{cc|cc|cc|cc|cc|cc}
inst & LB/UB & inst & LB/UB & inst & LB/UB & inst & LB/UB & inst & LB/UB & inst & LB/UB\\
\hline
001 & 23 & 002 & 32 & 003 & 29 & 004 & 43/44 & 005 & 42 & 006 & 47\\
007 & 41 & 008 & 44 & 009 & 57/72 & 010 & 38 & 011 & 44 & 012 & 45\\
013 & 36 & 014 & 45 & 015 & 43 & 016 & 63 & 017 & 62 & 018 & 60\\
019 & 59 & 020 & 62 & 021 & 51 & 022 & 59 & 023 & 51 & 024 & 56\\
025 & 69/70 & 026 & 54 & 027 & 55 & 028 & 64 & 029 & 43 & 030 & 20\\
031 & 70 & 032 & 80 & 033 & 78 & 034 & 73 & 035 & 73/77 & 036 & 100/106\\
037 & 116/138 & 038 & 86 & 039 & 99/111 & 040 & 87/91 & 041 & 27 & 042 & 29\\
043 & 105 & 044 & 103 & 045 & 86/87 & 046 & 110/128 & 047 & 103/107 & 048 & 76/77\\
049 & 29 & 050 & 94/109 & 051 & 29 & 052 & 85 & 053 & 97/113 & 054 & 92/100\\
055 & 91/97
\end{tabular}
\end{table}

\begin{table} \small
\centering
\caption{Lower and upper bounds for \benchPackD.}
\label{tab:lub:packd}
\begin{tabular}{cc|cc|cc|cc|cc}
inst & LB/UB & inst & LB/UB & inst & LB/UB & inst & LB/UB & inst & LB/UB\\
\hline
001 & 612 & 002 & 745/747 & 003 & 624/625 & 004 & 1381 & 005 & 983\\
006 & 1119 & 007 & 1082 & 008 & 1274 & 009 & 1593/1951 & 010 & 1216\\
011 & 940 & 012 & 1234/1241 & 013 & 829 & 014 & 1565 & 015 & 1198\\
016 & 1783/1813 & 017 & 1641/1651 & 018 & 1462/1480 & 019 & 1526/1542 & 020 & 1661\\
021 & 1606 & 022 & 1787 & 023 & 1092 & 024 & 1625 & 025 & 2061/2147\\
026 & 926 & 027 & 1789/1793 & 028 & 1897/1962 & 029 & 1233 & 030 & 597\\
031 & 1949 & 032 & 2943 & 033 & 3390 & 034 & 2371 & 035 & 2305\\
036 & 2175/2191 & 037 & 3325/3614 & 038 & 2180 & 039 & 2730/2734 & 040 & 3024\\
041 & 679 & 042 & 838 & 043 & 2439 & 044 & 3050 & 045 & 2712\\
046 & 3243/3277 & 047 & 2740/2745 & 048 & 2446 & 049 & 675 & 050 & 2687/2716\\
051 & 838 & 052 & 2253 & 053 & 2521 & 054 & 2750 & 055 & 2628
\end{tabular}
\end{table}

\end{WithAppendix}

\end{document}